\documentclass[runningheads]{llncs}
\usepackage{epsfig}
\usepackage{graphicx}
\usepackage{amsmath}
\usepackage{amssymb}
\usepackage{gensymb}
\usepackage{multirow}
\usepackage{rotating}
\usepackage{fancyhdr}
\usepackage{latexsym}
\usepackage{url}
\usepackage{algorithm}
\usepackage{algorithmic}
\usepackage{bm}
\usepackage{stmaryrd}
\usepackage{dsfont}
\usepackage{epstopdf}
\usepackage{booktabs}
\usepackage{dashrule}

\AppendGraphicsExtensions{.tiff}

\usepackage{caption}
\usepackage{subcaption}

\usepackage[pagebackref=true,breaklinks=true,letterpaper=true,colorlinks,bookmarks=false]{hyperref}

\usepackage[width=122mm,left=12mm,paperwidth=146mm,height=193mm,top=12mm,paperheight=217mm]{geometry}

\index{Duc Minh, Vo}

\begin{document}
\addtolength{\baselineskip}{-0.25pt}
\pagestyle{headings}
\mainmatter
\title{Visual-Relation Conscious Image Generation from Structured-Text\thanks{\scriptsize The authors are thankful to Zehra Hay{\i}rc{\i} for her valuable comments on this work.}}

\titlerunning{Visual-Relation Conscious Image Generation from Structured-Text} 

\author{Duc Minh Vo\inst{1} \and
Akihiro Sugimoto\inst{2}}

\authorrunning{Duc M. Vo and A. Sugimoto} 

\institute{Department of Informatics, \\ The Graduate University for Advanced Studies, SOKENDAI, Tokyo, Japan \and
National Institute of Informatics, Tokyo, Japan\\
\email{\{vmduc, sugimoto\}@nii.ac.jp}}

\maketitle

\begin{figure}[!h]
	\centering
	\includegraphics[width=0.95\linewidth]{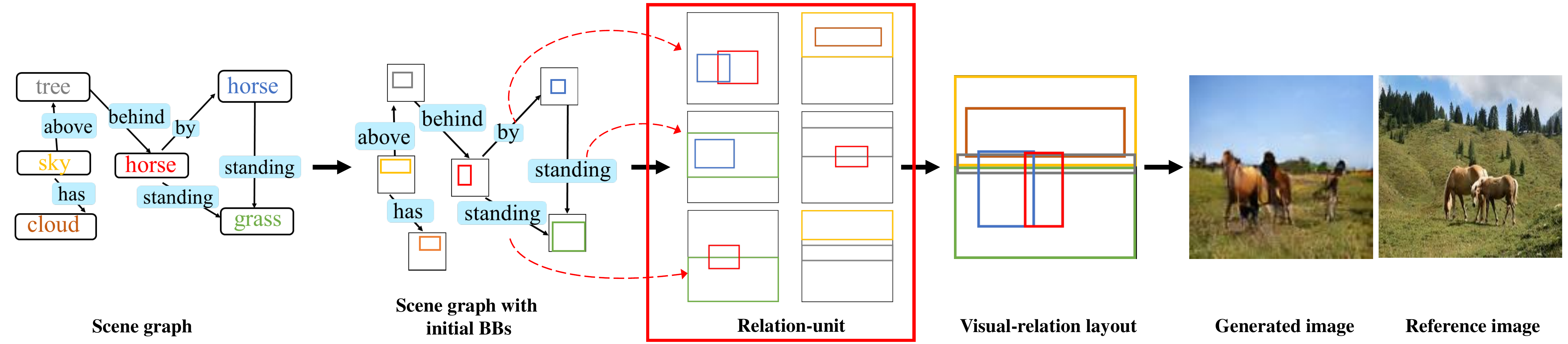}
	\caption{Overall framework of our proposed method. Given a structured-text (scene graph), our method firstly predicts initial bounding-boxes for entities using all available relations together, next takes individual relation one by one to infer a relation-unit for the relation, then unifies all the relation-units to produce visual-relation layout. Finally, the visual-relation layout is converted to image ($256 \times 256)$. Color of each entity bounding-box corresponds to that in the scene graph. Dotted arrow in red illustrates the individual usage of relations.} 
	\label{fig:example}
	\vspace*{-2\baselineskip}
\end{figure}

\begin{abstract}
We propose an end-to-end network for image generation from given structured-text that consists of the visual-relation layout module and the pyramid of GANs, namely stacking-GANs.
Our visual-relation layout module uses relations among entities in the structured-text in two ways: comprehensive usage and individual usage.
We comprehensively use all available relations together to localize initial bounding-boxes of all the entities. 
We also use individual relation separately to predict from the initial bounding-boxes relation-units for all the relations in the input text. 
We then unify all the relation-units to produce the visual-relation layout, i.e., bounding-boxes for all the entities so that each of them uniquely corresponds to each entity while keeping its involved relations.
Our visual-relation layout reflects the scene structure given in the input text.  
The stacking-GANs is the stack of three GANs conditioned on the visual-relation layout and the output of previous GAN, consistently capturing the scene structure. 
Our network realistically renders entities' details in high resolution while keeping the scene structure.
Experimental results on two public datasets show outperformances of our method against state-of-the-art methods.
\end{abstract}

\section{Introduction}

Generating photo-realistic images from text descriptions (T2I) is one of the major problems in computer vision.
Besides having a wide range of applications 
such as intelligent image manipulation, it drives research progress in multimodal learning and inference across vision and language~\cite{johnson2018image,johnson2015image,li2017msdn}.

The GANs~\cite{ian2014generative} conditioned on unstructured text description~\cite{reed2016generative,han2017stackgan,reed2016learning,dong2017semantic,Tao18attngan} show remarkable results in T2I. 
Stacking such conditional GANs has shown even more ability of progressively rendering a more and more detailed entity in high resolution~\cite{han2017stackgan,Tao18attngan}.
However, in more complex scenarios where sentences are with many entities and relations, their performance is degraded.
This is because they use only entity information in given text descriptions for rendering a specific entity, leading to a poor layout of multiple entities in generated images.

In the presence of multiple entities, besides the details of each entity, how to localize all the entities so that they reflect given relations becomes crucial for better image generation.
Indeed, recent work~\cite{johnson2018image,li2019pastegan,ashual2019scenegeneration,hong2018inferring} show the effectiveness of inferring the scene layout first from given text descriptions.
Johnson+\cite{johnson2018image}, Li+\cite{li2019pastegan}, and Ashual+\cite{ashual2019scenegeneration} use structured-text, i.e., scene graphs~\cite{johnson2015image}, first to construct a scene layout by predicting bounding boxes and segmentation masks for all entities, then convert it to an image.
Hong+\cite{hong2018inferring} constructs a semantic layout, a scene structure based on object instances, from input text descriptions and converts the layout into an image.
However, those mentioned methods~\cite{johnson2018image,li2019pastegan,ashual2019scenegeneration,hong2018inferring} 
aggregate all relations in which each entity is involved, and then localize all entities' bounding-boxes at the same time.
As a result, the predicted bounding-boxes do not preserve the relations among entities well.
Localizing entities faithfully by preserving their relations given in text descriptions is desired.

We leverage advantages of the pyramid of GANs and inferring the scene layout, proposing a GAN-based model for T2I 
where our network steps further in relation usage by employing not only all available relations together
but also individual relation separately.
We refer the former usage of relations as \textit{comprehensive} while the latter as \textit{individual}.  
Our network has two steps: (1) inferring from input the \textit{visual-relation layout}, i.e., 
localized bounding-boxes for all the entities so that each of which uniquely corresponds to each entity and faithfully preserves relations between the entities,
and (2) progressively generating coarse-to-fine images with the pyramid of GANs, namely stacking-GANs, conditioned on the visual-relation layout.
The first step takes the comprehensive usage of relations first to generate initial bounding-boxes (BBs) for entities as in~\cite{johnson2018image,li2019pastegan,ashual2019scenegeneration,hong2018inferring}, and then takes the individual usage to predict a relation-unit for each \textit{subject--predicate--object} relation 
where all the relations in the input are extracted through its scene graph~\cite{johnson2015image}.
Each relation-unit consists of \textit{two} BBs that participate in the relation: one for a ``subject" entity and one for an ``object" entity.  
Since one entity may participate in multiple relations, we then unify all the relation-units into refined (entity) BBs (including their locations and sizes) so that each of them uniquely corresponds to one entity while keeping their relations in the input text.
Aggregating the refined BBs allows us to infer the visual-relation layout reflecting the scene structure given in the text.
In the second step, three GANs progressively generate images where entities are rendered in more and more details while preserving the scene structure.
At each level, a GAN is conditioned on the visual-relation layout and the output of previous GAN.
Our network is trained in a fully end-to-end fashion.

The main contribution of our proposed method is our introduction to the individual usage of \textit{subject--predicate--object} relations 
for localizing entity bounding-boxes, so that our proposed \textit{visual-relation layout} surely preserves the visual relations among entities.
In addition, we stack and condition GANs on the visual-relation layout to progressively render realistic detailed entities that keep their relations even from complex text descriptions.
Experimental results on COCO-stuff~\cite{caesar2018coco} and GENOME~\cite{krishna2017visualgenome} demonstrate outperformances of our method against state-of-the-arts.
Fig.~\ref{fig:example} shows the overall framework of our proposed method.

\section{Related work}

Recent GAN-based methods have shown promising results on T2I~\cite{johnson2018image,reed2016generative,han2017stackgan,dong2017semantic,Tao18attngan,hong2018inferring,reed2016nips}.
They, however, struggle to faithfully reproduce complex sentences with many entities and relations because of the gap between text and image representations.

To overcome the limitation of GANs conditioned on text descriptions, a two-step approach was proposed where inference of the scene layout as an intermediate representation between text and image is followed by using the layout to generate images~\cite{johnson2018image,li2019pastegan,ashual2019scenegeneration,hong2018inferring}.
Since the gap between the intermediate representation and image is smaller than that of text and image, this approach generates more realistic images.
Zhao+\cite{zhaobo2019layout2im} and Sun+\cite{sun2019image} propose a combination of ground-truth (GT) layout and entity embeddings to generate images. 
Hong+\cite{hong2018inferring} infers a scene layout by feeding text descriptions into a LSTM.  
More precisely, they use a LSTM to predict BBs for all entities independently, then employ a bi-directional conv-LSTM to generate entity shapes from each predicted BB without using any relation. 
The function of the bi-directional conv-LSTM used here is just the putting-together.
They then combine the layout with text embeddings obtained from the pre-trained text encoder~\cite{reed2016learning}, and use a cascade refinement network (CRN)~\cite{chen2017photographic} for generating images.

Johnson+\cite{johnson2018image}, Li+\cite{li2019pastegan}, and Ashual+\cite{ashual2019scenegeneration} employ a scene graph~\cite{johnson2015image} to predict a layout and then condition CRN~\cite{chen2017photographic} on the layout.
The graph convolution network (GCN) used in these methods aggregates available relations of all the entities together along the edges of the scene graph.
Namely, only the comprehensive usage of relations is employed.
Consequently, individual relation information is lost at the end of GCN because of the averaging operation on entity embeddings.
Averaging entity embeddings means mixing different relations in which a single entity is involved, resulting in failure of retaining individual relation information.
Different from~\cite{johnson2018image}, \cite{li2019pastegan} retrieves entity appearances from a pre-defined tank while \cite{ashual2019scenegeneration} adds entity appearances to the layout before feeding it to the generation part. 
The layout in ~\cite{johnson2018image,li2019pastegan,ashual2019scenegeneration,hong2018inferring} is constructed through only the comprehensive usage of relation among entities for BBs' localization,
leading poor scene structure as a whole even if each entity is realistically rendered.

Our main difference from the aforementioned methods is to construct the visual-relation layout using
\textit{subject--predicate--object} relations between entities extracted from an input structured-text not only comprehensively but also individually.
Recursively conditioning stacking-GANs on our constructed visual-relation layout enables us to progressively generate coarse-to-fine images that consistently preserve the scene structure given in texts.

\section{Proposed method}

\begin{figure}[t]
	\centering
	\includegraphics[width=\linewidth]{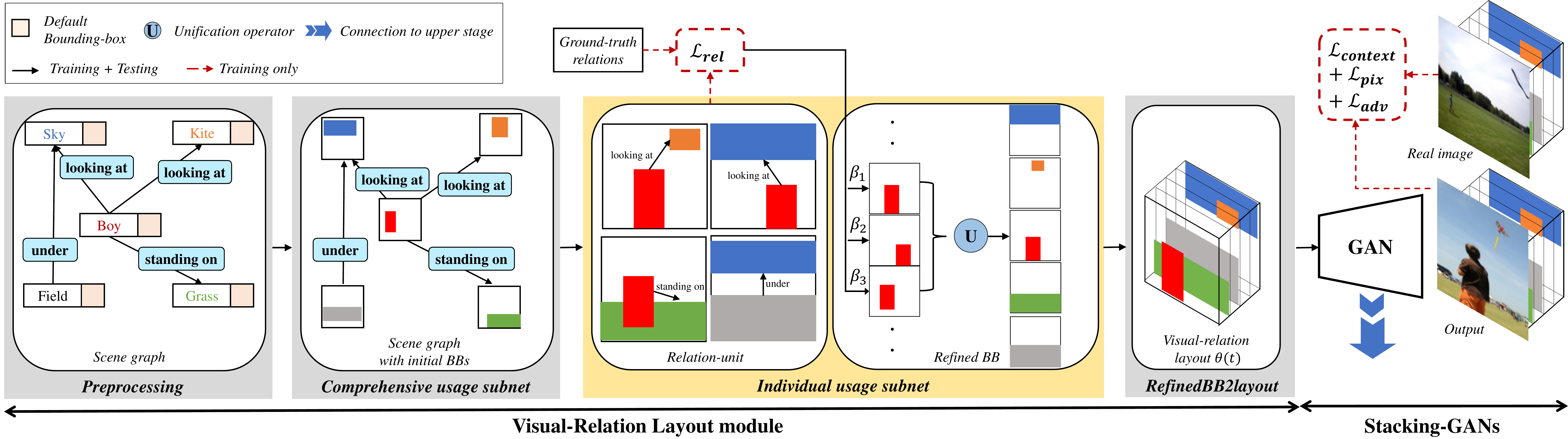}
    \vspace*{-0.5\baselineskip}
	\caption{Our proposed network model consisting of the visual-relation layout module and the Stacking-GANs.} 
	\label{fig:framework}
	\vspace*{-\baselineskip}
\end{figure}

Our method is decomposed into two steps: (1) inferring the visual-relation layout $\theta(t)$ from structured-text description $t$, and (2) generating an image from the visual-relation layout, namely $\hat{I} = G(\theta(t))$. 
To this end, we design an end-to-end network with two modules: the visual-relation layout module and the stacking-GANs (Fig.~\ref{fig:framework}). We train the network in a fully end-to-end manner.

\subsection{Visual-relation layout module}

The individual usage of relations in this module is inspired by the observation that individual relation information is lost after the comprehensive usage of relations~\cite{johnson2018image,li2019pastegan,ashual2019scenegeneration}.
Two way usage of relations leads to two different designs: (i) individual--comprehensive and (ii) comprehensive--individual.
The former design again will fail in retaining individual relation information,
resulting in poor layout like~\cite{johnson2018image,li2019pastegan,ashual2019scenegeneration}.
In contrast, the latter one is more promising because it is reasonable to initialize all entities' locations first and then to adjust them to meet their involved each relation.
We thus employ the comprehensive--individual design.

\begin{figure}[tb]
	\centering
	\includegraphics[width=0.95\linewidth]{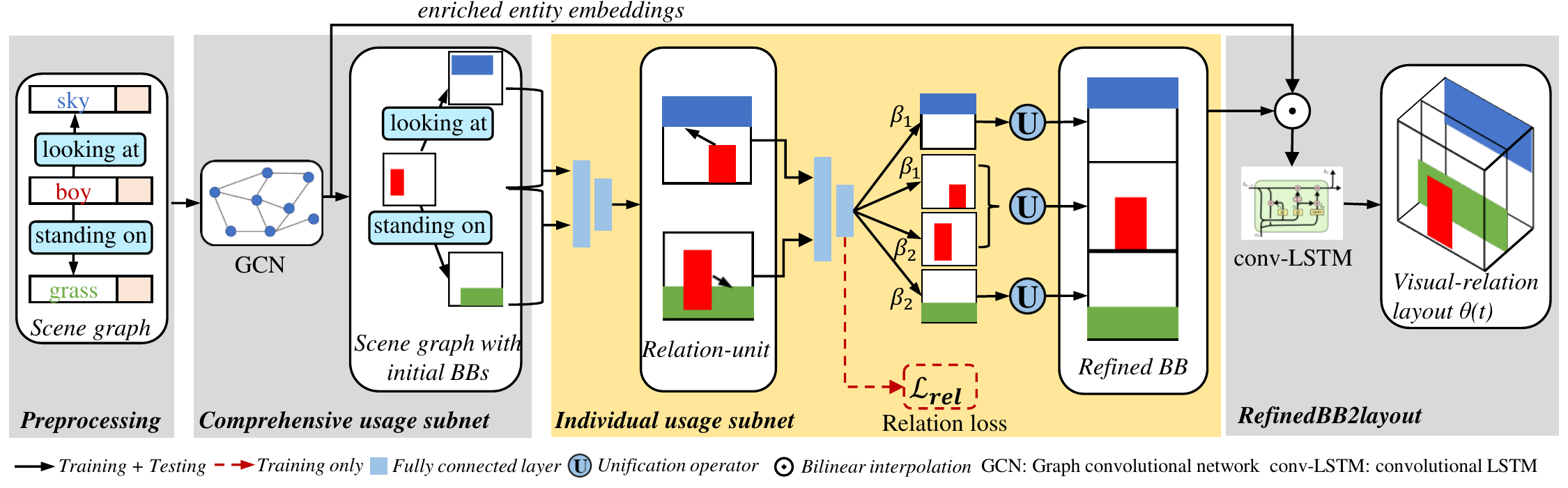}
    \vspace*{-0.5\baselineskip}
	\caption{Details of visual-relation layout module. This figure illustrates the prediction for two \textit{subject--predicate--object} relations.
	} 
	\label{fig:layout_module}
	\vspace*{-1.5\baselineskip}
\end{figure}

The visual-relation layout module constructs the visual-relation layout $\theta(t)$ from a given structured-text description $t$ (Fig.~\ref{fig:layout_module})
where $t$ is assumed to be converted into a scene graph~\cite{johnson2015image}, i.e., the collection of \textit{subject--predicate--object}'s.
After the pre-processing on converting $t$ to its scene graph,  
the comprehensive usage subnet in this module predicts initial BBs for all the entities in $t$ by aggregating all available relations together through GCN (``comprehensive usage").
The individual usage subnet takes each \textit{subject--predicate--object} relation from the scene graph one by one and
select the pair of initial BBs involved in the relation (predicate): one for ``subject" entity and one for ``object" entity. 
The subnet then adjusts the location and size of the pair of initial BBs using the relation (``individual usage") to have a relation-unit for the relation.
Since one entity may participate in multiple relations, it next unifies relation-units so that each entity uniquely has a single BB (called refined BB) that is further adjusted in location and size using weights learned from all the participating relations.
The RefinedBB2layout subnet constructs the visual-relation layout by aggregating all the refined BBs together using conv-LSTM.

\noindent
\textbf{Preprocessing.}
Similar to~\cite{johnson2018image}, we convert the structured-text $t$ to its scene graph $(E, P)$ where $E \subseteq \mathcal{C}$ 
and $P \subseteq \mathcal{C}\times\mathcal{R}\times\mathcal{C}$.
$\mathcal{C}$ and $\mathcal{R}$ are the set of categories and the set of relations given in a dataset.
An edge of $(E,P)$ is associated with one \textit{subject--predicate--object}.
It is directed and represented by $(e^{\rm s}, p, e^{\rm o})$ with entities $e^{\rm s}, e^{\rm o} \in E$ and predicate $p \in \mathcal{R}$ ($\rm s$ and $\rm o$ indicate subject and object).

Like~\cite{johnson2018image}, we use a learned embedding layer to produce the entity embedding with the size of $1 \times |\mathcal{C}|$
and the predicate embedding with the size of $1 \times |\mathcal{R}|$ for any of all the entities and predicates appearing in $(E,P)$.
Any entity embedding is associated with a single default BB presented by $[x, y, w, h] \in [0,1]^4$ where $x$ is the \textit{left coordinate}, $y$ is the \textit{top coordinate}, $w$ is the \textit{width}, and $h$ is the \textit{height}. We set $x = y = 0$ and $w = h = 1$ as default. 
This process ensures that all the entities appear in the image.
In practice, we concatenate the default BB and its associated entity embedding to produce the vector with the size of $1 \times (|\mathcal{C}| + 4)$.

\noindent
\textbf{Comprehensive usage subnet.}
This subnet applies the comprehensive usage to predict a single initial BB for each entity appearing in $t$
as in~\cite{johnson2018image,li2019pastegan,ashual2019scenegeneration,hong2018inferring}.
This subnet gives us initial locations and sizes of entities and they do not necessarily satisfy the relations given in $t$.

In order to aggregate all information along the edges in the scene graph, we employ GCN~\cite{johnson2018image}. 
Our GCN is mostly identical to~\cite{johnson2018image} with a modification that produces 388 outputs instead of 384 not only to enrich entity/predicate embeddings as in~\cite{johnson2018image,li2019pastegan,ashual2019scenegeneration} but also to infer initial BBs.
We do not use the average pooling layer on top of GCN to retain individual relation information.

For each edge $k$ of $(E, P)$, the triplet $(\bm{e}^{\rm s}_{k}, \bm{p}_{k}, \bm{e}^{\rm o}_{k})$ and two default BBs with the size of $1 \times (|\mathcal{C}| + |\mathcal{R}| + |\mathcal{C}| + 8)$ are processed to give 
enriched $\bm{e}'^{\rm s}_{k}$, $\bm{p}'_{k}$, and $\bm{e}'^{\rm o}_{k}$ embeddings with the size of $1 \times 128$ each, separately, and a pair of initial BBs (one for ``subject" and one for ``object") with the size of $1 \times 4$ each.

\noindent
\textbf{Individual usage subnet.}
Since the initial BBs of the entities do not always satisfy the relations given in $t$, 
we adjust their locations and sizes using each relation separately. 
For each relation, we select a pair of initial BBs corresponding to the ``subject" and ``object" involved in the relation, and
adjust the locations and sizes of the pair of BBs using the relation to have a relation-unit for the relation consisting of \textit{two} BBs for ``subject" and ``object" entities in the relation.
This process causes the situation where multiple BBs correspond to the same entity, as different relations may involve same entities in common. 
We thus move to focus on each entity to unify its corresponding BBs into a single BB (called refined BB) where we use weights learned to retain all the relations.
Accordingly, the function of this subnet is two-fold: relation-unit prediction using individual relation separately and unification of multiple BBs corresponding to the same entity into a single refined BB.
The subnet is built upon two fully-connected layers followed by a ReLU layer~\cite{Nair2010Rectified} producing 512 and 8 outputs.

For each edge $k$ of scene graph $(E, P)$, its enriched embeddings and its corresponding pair of initial BBs with the size of $1 \times 392(=128 + 4 + 128 + 128 + 4)$ are fed into this subnet to infer relation-unit $(\bm{b}_{k}^{\rm s}, \bm{b}_{k}^{\rm o})$ with the size of $1 \times 8$.
Each BB ($\bm{b}_{k}^{\rm s}$ or $\bm{b}_{k}^{\rm o}$) in the relation-unit is associated with enriched embedding either $\bm{e}'^{\rm s}_{k}$ or $\bm{e}'^{\rm o}_{k}$, respectively for ``subject" or ``object".
We remark that the total number of obtained BBs is $|\{\bm{b}_{k}^{\rm s}, \bm{b}_k^{\rm o}\}| = 2 \times |P|$, which is in general larger than $|E|$.

To encourage the refined BB of each entity to keep its involved relations, we use the relation loss $\mathcal{L}_{rel}$ (Sec. \ref{loss function}) in a supervised manner.
This is because $\mathcal{L}_{rel}$ indicates the degree of retaining the involved relations in terms of relation-unit.

For entity $e_i \in E$, let $\bm{B}_{i}=\{\bm{B}_{i\nu}\}$ denote the set of its corresponding BBs (appearing in different relation-units) and $\beta_{i}=\{\beta_{i\nu}\}$ be the set of their weights.
We define the refined BB: 
$\bm{\hat{B}}_{i} = \frac{{\sum_{\nu=1}^{|\bm{B}_{i}|} \{(1 + \beta_{i\nu}) \times \bm{B}_{i\nu}\}}}{{\sum_{\nu=1}^{|\bm{B}_{i}|} (1 + \beta_{i\nu})}}$.

Each weight in $\beta_{i}$ is obtained from the outputs of the softmax function in the relation auxiliary classifier using the relation loss $\mathcal{L}_{\rm rel}$.

At the beginning of training, relation-units cannot exactly reproduce their involved relations. 
Their weights thus tend to be close to $zero$, leading 
$\bm{\hat{B}}_{i}$ above almost similar to the simple average.
Our refined BBs may be close to those of~\cite{johnson2018image,li2019pastegan,ashual2019scenegeneration} at the beginning of training yet they keep their relations thanks to their weights.
As training proceeds, the contribution of the relation-units retaining relations consistent with text $t$ to the refined BB gradually increases.
As a result, the location and size of the refined BB are continuously altered to keep relations consistent with $t$.

For entity $e_i$, its embeddings that are associated with $\{\bm{B}_{i\nu}\}$'s over $\nu$ are averaged.
In this way, we obtain the set of refined BBs $\{\bm{\hat{B}}_i\}$ and their associated embeddings for all the entities in $E$. 
We remark that $|\{\bm{\hat{B}}_i\}| = |E|$.

If all the initial BBs completely keep their relations, the individual usage subnet works as the averaging operator as in~\cite{johnson2018image,li2019pastegan,ashual2019scenegeneration} and our visual-relation layout is similar to the layout by~\cite{johnson2018image,li2019pastegan,ashual2019scenegeneration}.
In practice, however, the comprehensive usage of relations cannot guarantee to completely keep the relations.
Our individual usage subnet plays the role of adjusting all the BBs in location and size to keep their relations as much as possible using each relation separately.

\noindent
\textbf{RefinedBB2layout subnet.}
In order to construct the visual-relation layout, we aggregate all the refined BBs and transfer them from the bounding-box domain to the image domain. 
This process should meet two requirements: (i) each entity in the image should be localized and resized to match its individual refined BB, and (ii) each entity should appear in the image even if some refined BBs overlap with each other.
To this end, we design \textit{refinedBB2layout} subnet as a learnable network rather than the putting-together operation. 
We build this subnet using a conv-LSTM~\cite{shi2015convolutional} with the 5 hidden states each outputting 128 channels. 

For $\bm{\hat{B}}_{i}$ of entity $e_i$, we first convert it to the binary mask with the size of $64 \times 64 \times 128$ whose element is 1 if and only if it is contained in $\bm{\hat{B}}_{i}$,
0 otherwise.
Then, we reshape its associated embedding from $1 \times 128$ to $1 \times 1 \times 128$.
Finally, the reshaped embedding is wraped to $\bm{\hat{B}}_{i}$ using the bilinear interpolation~\cite{jaderberg2015spatial} for the layout of entity $e_{i}$ ($64 \times 64 \times 128$).
To produce $\theta(t)$, we feed the sequence of entity layouts into the \textit{refinedBB2layout} subnet.
The size of $\theta(t)$ is $64 \times 64 \times 128$.

\subsection{Stacking-GANs}

We condition three GANs, namely stacking-GANs, on $\theta(t)$ to progressively generate coarse-to-fine images with the size of $n \times n \times 3$ ($n$ = 64, 128, 256).
Each GAN is identical to CRN~\cite{chen2017photographic}. 
Parameters are not shared by any GANs.

The first GAN generator receives the layout $\theta(t)$ and a standard Gaussian distribution noise as input while the others receive the bilinear upsampled~\cite{jaderberg2015spatial} layout $\theta(t)$ and the output of the last refinement layer from the previous GAN.
The discriminators receive an image-layout pair as their inputs. 
Each pair is either a real sample or a fake sample. 
A real sample consists of a real image and a real layout while
a fake one consists of a predicted layout and a generated or real image. 
These samples not only encourage the GAN to improve the quality of generated images but also give the helpful feedback to the layout predictor.

\subsection{Loss function} \label{loss function}

\noindent
\textbf{Relation loss} $\mathcal{L}_{\rm rel}$ is a cross entropy between relation-units and their GT relations
that is obtained by a relation auxiliary classifier.
The classifier is built upon two fully-connected layers producing 512 and $|\mathcal{R}|$ outputs. 
The first layer is followed by a ReLU layer while the second one ends with the $softmax$ function.

For each edge $k$ of $(E,P)$, its relation-unit and involved embeddings, i.e., $\bm{e}'^{\rm s}_{k}$, $\bm{b}_{k}^{\rm s}$, $\bm{e}'^{\rm o}_{k}$, and $\bm{b}_{k}^{\rm o}$, are concatenated in this order to have an input vector of $1 \times 264$.
We then feed this vector into the relation auxiliary classifier to obtain the probability distribution $\bm{w}_{k}$ of the relations over $\mathcal{R}$. $\bm{w}_{k}$ is a vector of $1 \times |\mathcal{R}|$ and contains all the predicates $p_k$ $\in \mathcal{R}$. 
We first obtain the \textit{index} of predicate $p_k$ $\in \mathcal{R}$. Since the order of predicates in $\bm{w}_{k}$ is the same as that in $\mathcal{R}$, the value at \textit{index} in $\bm{w}_{k}$ is the weight of $p_k$, which is used as the weight of the relation-unit ($\bm{b}_{k}^{\rm s}, \bm{b}_{k}^{\rm o}$) in the individual usage subnet.
Note that the weight of a relation-unit is used for the weight of both $\bm{b}_{k}^{\rm s}$ and $\bm{b}_{k}^{\rm o}$ involved in the relation-unit.

The relation loss is defined as:
$\mathcal{L}_{\rm rel} = -\sum_{k=1}^{|P|}\sum_{\nu'=1}^{|\mathcal{R}|}\bm{p}_{k}[\nu']\log(\bm{w}_{k}[\nu'])$. 
Minimizing the relation loss encourages relation-units to adjust their locations and sizes to meet the ``predicate" relation.  This is because the relation reflects the relative spatial locations among its associated relation-units.

\noindent
\textbf{Pixel loss:} $\mathcal{L}_{\rm pix} = || I - \hat{I} ||_2$, where $I$ is the ground-truth image and $\hat{I}$ is a generated image. 
The $\mathcal{L}_{\rm pix}$ is useful for keeping the quality of generated images.

\noindent
\textbf{Contextual loss}~\cite{mechrez2018contextual}:
 $\mathcal{L}_{\rm context}=-\log (CX(\Phi^l(I),\Phi^l(\hat{I})))$,
where $\Phi^l(\cdot)$ denotes the feature map extracted from layer $l$ of perceptual network $\Phi$, and $CX(\cdot)$ is the function that computes the similarity between image features.
$\mathcal{L}_{\rm context}$ is used to learn the context of an image since refined BBs may lose the context such as missing pixel information or the size of entity.

\noindent
\textbf{Adversarial loss}~\cite{ian2014generative} $\mathcal{L}_{\rm adv}$ encourages the stacking-GANs to generate realistic images. Since the discriminator also receives the real/predicted layout as its input, the $\mathcal{L}_{\rm adv}$ is helpful in training the visual-relation layout module as well.

In summary, we jointly train our network in an end-to-end manner to minimize:
$\mathcal{L} = \lambda_1\mathcal{L}_{\rm rel} + \lambda_2\mathcal{L}_{\rm pix} + \lambda_3\mathcal{L}_{\rm context} + \sum_{i=1}^{3}\lambda_4\mathcal{L}_{{\rm adv}i}$,
where $\lambda_{i}$ are hyper-parameters.
We compute $\mathcal{L}_{\rm adv}$ at each level in the stacking-GANs, while $\mathcal{L}_{\rm pix}$ and $\mathcal{L}_{\rm context}$ are computed at the third GAN.

\section{Experiments} \label{experiments}

\subsection{Dataset and compared methods}
\noindent
\textbf{Dataset.} We conducted experiments on challenging COCO-stuff~\cite{caesar2018coco} and Visual GENOME~\cite{krishna2017visualgenome} datasets, which have complex descriptions with many entities and relations in diverse context.
We followed~\cite{johnson2018image} to pre-process all the datasets: $|\mathcal{C}|=171$ and $|\mathcal{R}|=6$ 
(COCO-stuff~\cite{caesar2018coco}), and $|\mathcal{C}|=178$ and $|\mathcal{R}|=45$ (GENOME~\cite{krishna2017visualgenome}).

\noindent
\textbf{Compared methods.} We employed Johnson+\cite{johnson2018image} as the baseline ($64 \times 64$). 
To factor out the influence of image generator, we replaced the CRN in \cite{johnson2018image} by our stacking-GANs to produce higher resolution images ($128 \times 128$ and $256 \times 256$).
We also compared our method with Hong+\cite{hong2018inferring}, Zhang+\cite{han2017stackgan}, Xu+\cite{Tao18attngan}, Li+\cite{li2019pastegan}, Ashual+\cite{ashual2019scenegeneration}, Zhao+\cite{zhaobo2019layout2im}, and Sun+\cite{sun2019image}.
We reported the results in the original papers whenever possible.
For the methods that released at least one reference pre-trained model (\cite{Zhang} and \cite{Tao}),
we trained authors' provided codes (Zhang+\cite{han2017stackgan} and Xu+\cite{Tao18attngan}) on GENOME dataset.

\noindent
\textbf{Evaluation metrics.}
We use the inception score (IS)~\cite{salimans2016improved}, and Fr\'echet inception distance (FID)~\cite{Heusel17GANs} to evaluate the overall quality of generated images (implemented in~\cite{IS,FID}). 
We also use four metrics to evaluate the layout: the entity recall at IoU threshold ($R@\tau$), the relation IoU ($rIoU$), the relation score ($RS$)~\cite{tripathi2019using}, and the BB coverage.
Furthermore, to evaluate the relevance of generated images and input text descriptions, we use the image caption metrics: $BLEU$~\cite{papineni2002bleu}, $METEOR$~\cite{lavie05meteor}, and $CIDEr$~\cite{vedantam2015cider}.
We also evaluate the diversity of generated images using the diversity score~\cite{zhang2018perceptual} (implemented in~\cite{DS}).

To evaluate how much the predicted layout is consistent with the ground-truth (GT), we measure the agreement in size and location between predicted (i.e., refined) and GT BBs using the entity recall at IoU threshold: $R@\tau=|\{i \mid IoU(\bm{\hat{B}}_i,\bm{GT}_i) \geq \tau \}|/{N}$, where $\bm{\hat{B}}_i$ and $\bm{GT}_i$ are predicted and GT BBs for entity $e_i$, $N = \min (|\{\bm{\hat{B}}_i\}|, |\{\bm{GT}_i\}|)$ (we always observed $|\{\bm{\hat{B}}_i\}|=|\{\bm{GT}_i\}|$), $\tau$ is a IoU threshold, and $IoU(\cdot)$ denotes Intersection-over-Union metric.
Note that we used only the BBs that exist in both $\{\bm{\hat{B}}_i\}$ and $\{\bm{GT}_i\}$ to compute $R@\tau$.

We also evaluate the predicted layout using \textit{subject--predicate--object} relations.
For each \textit{subject--predicate--object} relation, we computed the IoU of the predicted ``subject" BB and its corresponding GT, and that for the ``object". We then multiplied the two IoUs to obtain the IoU for the relation.
$rIoU$ is the average over all the \textit{subject--predicate--object} relations.

We use the relation score (RS)~\cite{tripathi2019using} for COCO-stuff to evaluate the compliance of geometrical relation between predicted BBs. For each edge $k$ of scene graph $(E, P)$, we define $score(\bm{\hat{B}}_{k}^{\rm s}, \bm{\hat{B}}_k^{\rm o}) = 1$ if and only if the relative location between $\bm{\hat{B}}_{k}^{\rm s}$ and $\bm{\hat{B}}_{k}^{\rm o}$ satisfies the relation $p_k$, $0$ otherwise.
$RS = \sum_{k=1}^{|P|}score(\bm{\hat{B}}_{k}^{\rm s}, \bm{\hat{B}}_k^{\rm o}) / |P|$.

To evaluate how much BBs cover the area of the whole image, we compute the coverage of predicted BBs over the image area: $coverage=\bigcup_{i=1}^{|E|}\bm{\hat{B}}_i/({\rm \mbox{image area}})$.

We note that $R@\tau$ and $rIoU$ consider the consistency between predicted BBs and GT BBs, and $RS$ and $coverage$ are independent of GT BBs. In other words, $R@\tau$ and $rIoU$ evaluate absolute locations of BBs while $RS$ (and $coverage$ as well to some extent) does semantic relations. Therefore, they together effectively evaluate the layout in a wide range of aspects.

\subsection{Implementation and training details}

We optimized our model (built in PyTorch~\cite{PyTorch}) using the Adam optimizer with the recommended parameters~\cite{kingma2014auto} and the batch size of $16$ for $500$ epochs.
We used VGG-19~\cite{SimonyanZ14a} pre-trained on ImageNet 
as $\Phi$, and $l=conv4\_2$ to compute $\mathcal{L}_{\rm context}$.
Each model took about one week for training on a PC with GTX1080Ti $\times$ 2 while testing time was less than 0.5 second per structured-text input.

We trained the model except for the pre-processing in the end-to-end manner where 
we set $\lambda_1=\lambda_2=\lambda_3=\lambda_4=1$,
and do not pre-train each individual subset, meaning that we do not use any ground-truth BBs to train the visual-relation layout.
The layout predictor receives signals not only directly from the relation loss but also from the other losses. 
In an early stage of the training, the rendering part cannot generate reasonable images because the quality of BBs is poor. 
This means the signals from losses are strong, leading to quick convergence of the layout predictor.
As the training proceeds, the layout predictor properly works, and the rendering part gradually becomes better.
$\mathcal{L}_{\rm rel}$, at that time, keeps the layout predictor stable and more accurate.

\subsection{Comparison with state-of-the-arts}

\begin{figure}[!h]
    \centering
    \begin{subfigure}[b]{0.85\textwidth}
        \includegraphics[width=\textwidth]{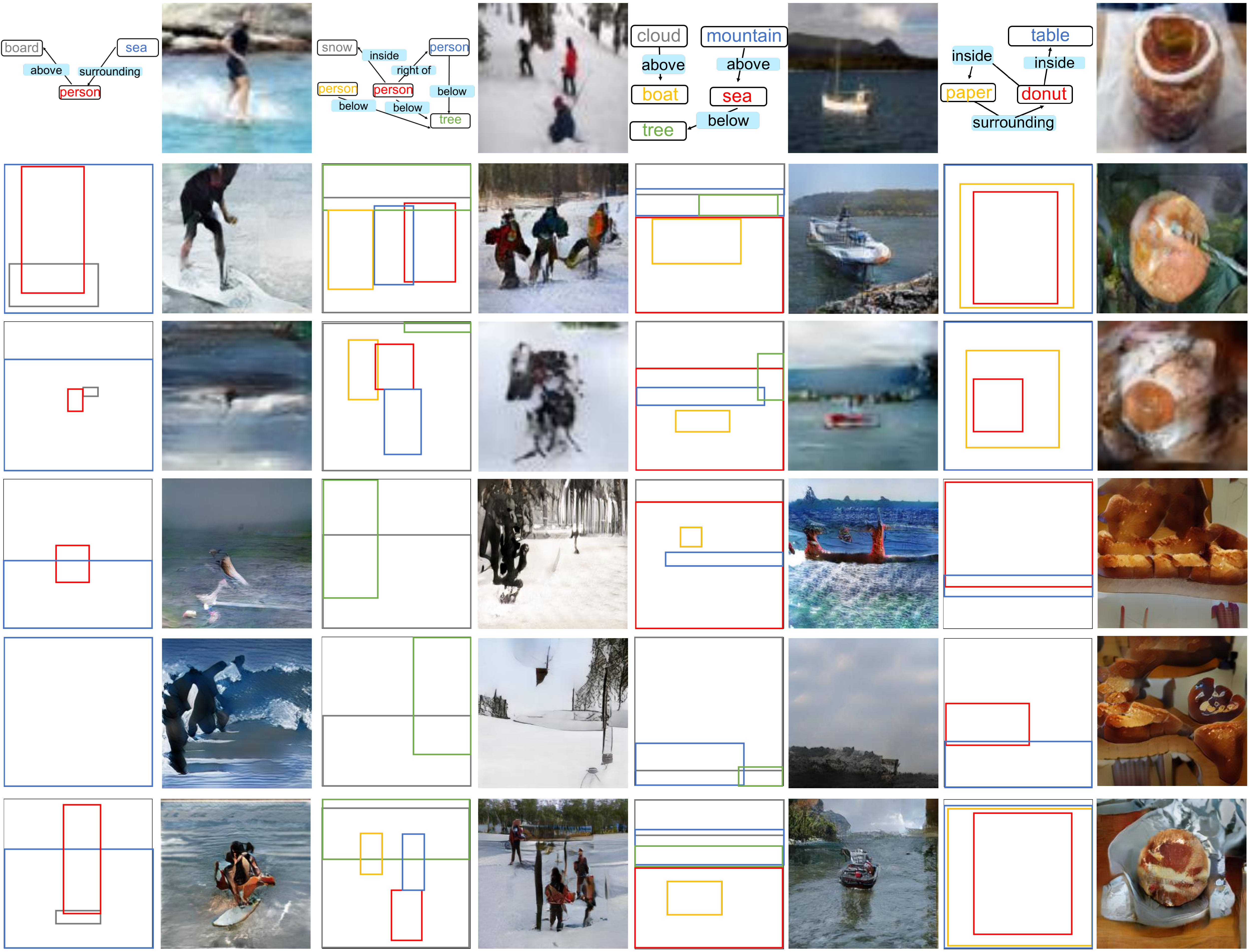}
        \caption{COCO-stuff dataset.}
        \label{fig:visualization_coco}
    \end{subfigure}
    \begin{subfigure}[b]{0.85\textwidth}
        \includegraphics[width=\textwidth]{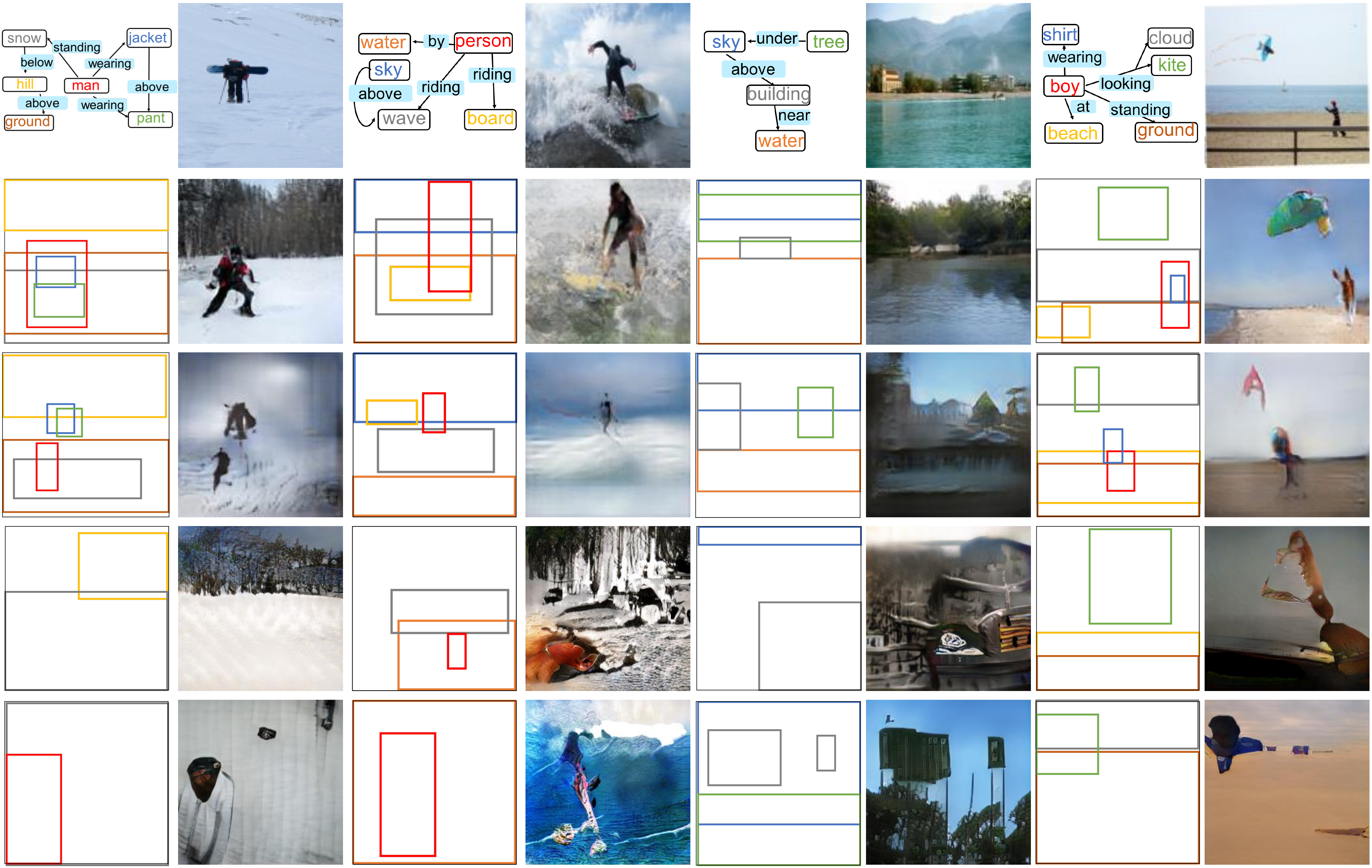}
        \caption{GENOME dataset.}
        \label{fig:visualization_genome}
    \end{subfigure}
    \vspace*{-0.5\baselineskip}
    \caption{Visual comparison on COCO-stuff~\cite{caesar2018coco},
	and GENOME~\cite{krishna2017visualgenome}. For each example, we show the scene graph and reference image at the first row. From second to the last rows, we show the layouts and images generated by our method ($256 \times 256$), Johnson+\cite{johnson2018image} ($64 \times 64$), Zhang+\cite{han2017stackgan} ($256 \times 256$), Xu+\cite{Tao18attngan} ($256 \times 256$), and Ashual+\cite{ashual2019scenegeneration} ($256 \times 256$, COCO-stuff only, GT layout). The color of each entity BB corresponds to that in the scene graph.
	Zoom in for best view.
	} 
    \label{fig:visual-comparison}
    \vspace*{-1.5\baselineskip}
\end{figure}

\noindent
\textbf{Qualitative evaluation.}
Fig.~\ref{fig:visual-comparison} shows examples of the results obtained by our method and SOTAs~\cite{johnson2018image,han2017stackgan,Tao18attngan,ashual2019scenegeneration} on COCO-stuff~\cite{caesar2018coco} and GENOME~\cite{krishna2017visualgenome} datasets.
It shows that the generated images by our method successfully preserve the scene structure given in text descriptions, indicating that our proposed visual-relation layouts are highly consistent with those of GTs.
We see that the results by Johnson+\cite{johnson2018image} have reasonable layouts,
however, their layouts failed to keep all relations well and the visual impression of their results is not good.
The results by Zhang+\cite{han2017stackgan} and Xu+\cite{Tao18attngan} are clear in (entities) details but they lose the scene structure (some entities disappear).
The results by Ashual+~\cite{ashual2019scenegeneration} (COCO-stuff only) are more impressive than ours to some extent, however, they use GT layout and pre-defined entities' appearances.

\noindent
\textbf{Quantitative evaluation.}
We classify all the compared methods into three: (A) Johnson+~\cite{johnson2018image}, Hong+\cite{hong2018inferring}, Li+\cite{li2019pastegan}, and Ashual+\cite{ashual2019scenegeneration} (which firstly infer a layout and then convert it to an image), (B) Zhang+\cite{han2017stackgan} and Xu+\cite{Tao18attngan} (which are directly conditioned on texts), and (C) Zhao+\cite{zhaobo2019layout2im} and Sun+\cite{sun2019image} (which are directly conditioned on ground-truth layouts).

Table~\ref{tab:quantitative_IS_FID} shows that our method (almost) outperforms (A) in $IS$ and $FID$ on both COCO-stuff and GENOME.
In comparison with (B), our method achieves the best in $FID$ on both the datasets, the best on GENOME and the second best on COCO-stuff in $IS$. 
Xu+\cite{Tao18attngan} achieves better $IS$ on COCO-stuff than us because (i) Xu+\cite{Tao18attngan} focuses on generating images in good human perception based on entity information and (ii) COCO-stuff has less complex relations, in other words, layouts may be less important.
On GENOME, however, text descriptions are more complex with many entities and relations, and their results are degraded 
due to poor layouts as seen later in Table~\ref{tab:quantitative_layout}.
Table~\ref{tab:quantitative_IS_FID} also shows that the scores of our completed model are comparable to those of (C), meaning that our (predicted) visual-relation layout is close to the GT layout. When replacing the predicted layout by the GT (the 17th row), our results achieve the same level with (C).
We may thus conclude that our method is more effective and promising than the others.

\begin{table}[tb]
\centering
\caption{Comparison of the overall quality using $IS$ and $FID$.
From the 4th to the 16th rows: group (A) and (B) (the best in \textcolor[rgb]{0,0,1}{blue}; the second best in \textcolor[rgb]{1,0,0}{red}).
From the 17th to the 19th rows: group (C) (\textbf{bold} indicates the best).
Scores inside the parentheses indicate those reported in the original papers.} 
\label{tab:quantitative_IS_FID}
\resizebox{\linewidth}{!}{
\begin{tabular}{l|@{\hspace*{0.35em}}c@{\hspace*{0.35em}}c@{\hspace*{0.35em}}c@{\hspace*{0.35em}}c@{\hspace*{0.35em}}c@{\hspace*{0.35em}}c@{\hspace*{0.35em}}c@{\hspace*{0.35em}}c@{\hspace*{0.35em}}c@{\hspace*{0.35em}}c@{\hspace*{0.35em}}c|@{\hspace*{0.35em}}c@{\hspace*{0.35em}}c@{\hspace*{0.35em}}c@{\hspace*{0.35em}}c@{\hspace*{0.35em}}c@{\hspace*{0.35em}}c@{\hspace*{0.35em}}c@{\hspace*{0.35em}}c@{\hspace*{0.35em}}c@{\hspace*{0.35em}}c@{\hspace*{0.35em}}c}
\midrule[0.1pt]
& \multicolumn{10}{c}{\textbf{IS} $\Uparrow$} & & \multicolumn{11}{c}{\textbf{FID} $\Downarrow$}\\
\cmidrule{2-23} 
{\textbf{Dataset}} &  \multicolumn{5}{c}{\textbf{COCO-stuff}~\cite{caesar2018coco}} & & & \multicolumn{3}{c}{\textbf{GENOME}~\cite{krishna2017visualgenome}} & & \multicolumn{5}{c}{\textbf{COCO-stuff}~\cite{caesar2018coco}} & & \multicolumn{5}{c}{\textbf{GENOME}~\cite{krishna2017visualgenome}}\\
\cmidrule{2-6} \cmidrule{8-12} \cmidrule{13-17} \cmidrule{19-23}
{\textbf{Image size}} & {$64 \times 64$} & & {$128 \times 128$} & & {$256 \times 256$} & & {$64 \times 64$} & & {$128 \times 128$} & & {$256 \times 256$} & {$64 \times 64$} & & {$128 \times 128$} & & {$256 \times 256$} & & {$64 \times 64$} & & {$128 \times 128$} & & {$256 \times 256$} \\
\midrule[0.1pt]
Ours w/o individual usage &  7.02$\pm$0.19  & & 8.12$\pm$0.41  & & 9.95$\pm$0.31 & & 5.48$\pm$0.16 & & 5.66$\pm$0.26 & & 5.91$\pm$0.41 & 63.28 & & 59.52 & & 55.21 & & 72.42 & & 72.02 & & 71.49 \\
Ours w/o weighted unification  & 7.10$\pm$0.27 & & 8.64$\pm$0.37 & & 10.49$\pm$0.41 & & 5.99$\pm$0.22 & & 6.61$\pm$0.31 & & 7.32$\pm$0.37 & 61.89 & & 57.20 & & 49.16 & & 69.37 & & 60.89 & & 57.18 \\
Ours w/o refinedBB2layout & 7.23$\pm$0.20 & & 8.70$\pm$0.35 & & 10.50$\pm$0.37 & & 6.11$\pm$0.25 & & 6.93$\pm$0.29 & & 7.87$\pm$0.33 & 57.68 & & 53.81 & & 46.55 & & 67.65 & & 58.54 & & 54.45 \\
\midrule[0.1pt]
Ours w/o $\mathcal{L}_{\rm pix}$ & 7.29$\pm$0.17 & & 9.26$\pm$0.31 & & 11.36$\pm$0.40 & & 6.05$\pm$0.15 & & 8.26$\pm$0.27 & & 8.66$\pm$0.36 & 56.81 & & 51.02 & & 43.18 & & 70.18 & & 60.02 & & 58.63 \\
Ours w/o $\mathcal{L}_{\rm context}$ & 7.56$\pm$0.11 & & 9.68$\pm$0.33 & & 11.47$\pm$0.42 & & 6.37$\pm$0.16 & & 8.41$\pm$0.22 & & 8.97$\pm$0.31 & 50.89 & & 47.22 & & 40.10 & & 68.20 & & 56.39 & & 53.75 \\
Ours w/o $\mathcal{L}_{\rm adv}$ & 7.31$\pm$0.19 & & 9.47$\pm$0.34 & & 11.41$\pm$0.47 & & 6.30$\pm$0.19 & & 8.39$\pm$0.20 & & 8.96$\pm$0.39 & 56.24 & & 50.87 & & 41.05 & & 68.34 & & 57.23 & & 53.86 \\
\textbf{Ours} (completed model) &  9.20$\pm$0.32 & & \textcolor[rgb]{1,0,0} {12.01$\pm$0.40} & & 14.20$\pm$0.45 & & \textcolor[rgb]{0,0,1} {7.97$\pm$0.30} & & \textcolor[rgb]{0,0,1} {9.24$\pm$0.41} & & \textcolor[rgb]{0,0,1} {11.75$\pm$0.43} & \textcolor[rgb]{0,0,1} {35.12} & & \textcolor[rgb]{0,0,1} {29.12} & & \textcolor[rgb]{0,0,1} {27.39} & & \textcolor[rgb]{0,0,1} {58.37} & & \textcolor[rgb]{0,0,1} {50.19} & & \textcolor[rgb]{0,0,1} {36.79} \\
\midrule[0.1pt]
Johnson+~\cite{johnson2018image} & (6.70$\pm$0.10) & & 7.13$\pm$0.24 & & 7.25$\pm$0.47 & &  (5.50$\pm$0.10) & & 5.72$\pm$0.33 & & 5.81$\pm$0.37 & 67.99 & & 65.23 & & 64.19 & & \textcolor[rgb]{1,0,0}{73.39} & & \textcolor[rgb]{1,0,0}{69.48} & & 68.42\\
Hong+~\cite{hong2018inferring} & --- & & (11.46$\pm$0.09) & & --- & &  --- & & --- & & --- & --- & & --- & &  --- & & --- & & --- & & --- \\
Li+~\cite{li2019pastegan} & \textcolor[rgb]{1,0,0}{(9.40$\pm$0.20)} & & --- & & --- & &  \textcolor[rgb]{1,0,0}{(7.30$\pm$0.20)} & & --- & & --- & --- & & --- & &  --- & & --- & & --- & & --- \\
Ashual+~\cite{ashual2019scenegeneration} & (7.90$\pm$0.20) & & (10.40$\pm$0.40) & & \textcolor[rgb]{1,0,0}{(14.50$\pm$0.70)} & &  --- & & --- & & --- & (65.30) & & (75.40) & &  (81.00) & & --- & & --- & & --- \\
\multicolumn{23}{c}{\dotfill} \\
Zhang+~\cite{han2017stackgan} & 7.79$\pm$0.32 & & 8.49$\pm$0.52 & & (10.62$\pm$0.19) & & 6.35$\pm$0.16 & & 6.44$\pm$0.25 & & 7.39$\pm$0.38 & 87.21 & & 85.37 & & 78.19 & & 108.68 & & 86.17 & & 77.95 \\
Xu+~\cite{Tao18attngan} & \textcolor[rgb]{0,0,1} {11.78$\pm$0.14} & & \textcolor[rgb]{0,0,1} {19.11$\pm$0.28} & & \textcolor[rgb]{0,0,1} {(25.89$\pm$0.47)} & & 6.38$\pm$0.22 & & \textcolor[rgb]{1,0,0}{6.88$\pm$0.32} & &  \textcolor[rgb]{1,0,0}{8.20$\pm$0.35} & \textcolor[rgb]{1,0,0}{50.06} & & \textcolor[rgb]{1,0,0}{43.98} & & \textcolor[rgb]{1,0,0}{34.48} & & 96.40 & & 83.39 & & \textcolor[rgb]{1,0,0}{72.11} \\
\multicolumn{23}{c}{\dotfill} \\
Ours with GT layout & \bf 10.36$\pm$0.41 & &  13.73$\pm$0.59 & & \bf 14.78$\pm$0.65 & & \bf 8.87$\pm$0.57 & &  10.04$\pm$0.45 & & \bf 12.03$\pm$0.37 & \bf 30.98 & & \bf 27.74 & & \bf 26.32 & &  45.63 & & 40.96 & & \bf 27.33 \\
Zhao+~\cite{zhaobo2019layout2im} (GT layout) & (9.10$\pm$0.10) & & --- & & --- & &  (8.10$\pm$0.10) & & --- & & --- & --- & & --- & &  --- & & --- & & --- & & --- \\
Sun+~\cite{sun2019image} (GT layout) & (9.80$\pm$0.20) & & \bf (13.80$\pm$0.40) & & --- & &  (8.70$\pm$0.40) & & \bf (11.10$\pm$0.60) & & --- & (34.31) & & (29.65) & &  --- & & \bf (34.75) & & \bf (29.36) & & --- \\
\midrule[0.1pt]
GT & 16.25$\pm$0.38 & & 25.89$\pm$0.47 & & 32.61$\pm$0.69 & & 13.92$\pm$0.42 & & 21.43$\pm$1.03 & & 31.22$\pm$0.65 & --- & & --- & &  --- & & --- & & --- & & -- \\
\midrule[0.1pt]
\end{tabular}
}
\vspace*{-1.75\baselineskip}
\end{table}

\begin{table}[tb]
\centering
\caption{Comparison of the scene structure using $R@\tau$, $rIoU$, $RS$, and $coverage$ (larger is better; the best in \textbf{bold}).}
\label{tab:quantitative_layout}
\resizebox{\linewidth}{!}{
\begin{tabular}{l|@{\hspace*{0.35em}}c@{\hspace*{0.35em}}c@{\hspace*{0.35em}}c@{\hspace*{0.35em}}c@{\hspace*{0.35em}}c@{\hspace*{0.35em}}c@{\hspace*{0.35em}}c@{\hspace*{0.35em}}c@{\hspace*{0.35em}}c@{\hspace*{0.35em}}c@{\hspace*{0.35em}}c@{\hspace*{0.35em}}c@{\hspace*{0.35em}}c@{\hspace*{0.35em}}c|@{\hspace*{0.35em}}c@{\hspace*{0.35em}}c@{\hspace*{0.35em}}c@{\hspace*{0.35em}}c@{\hspace*{0.35em}}c@{\hspace*{0.35em}}c@{\hspace*{0.35em}}cc@{\hspace*{0.35em}}c@{\hspace*{0.35em}}c@{\hspace*{0.35em}}c@{\hspace*{0.35em}}c}
\midrule[0.3pt]
{\textbf{Dataset}} &  \multicolumn{13}{c}{\textbf{COCO-stuff}~\cite{caesar2018coco}} & & \multicolumn{11}{c}{\textbf{GENOME}~\cite{johnson2015image}} \\
\cmidrule{2-26}
{\textbf{Metric}} &  \multicolumn{7}{c}{\textbf{$R@\tau$}} & & \textbf{$rIoU$} & & \textbf{$RS$} & & \textbf{$coverage$} & & \multicolumn{7}{c}{\textbf{$R@\tau$}} & & \textbf{$rIoU$} & & \textbf{$coverage$}\\
\cmidrule{2-8} \cmidrule{10-10} \cmidrule{12-12} \cmidrule{14-14} \cmidrule{16-22} \cmidrule{24-24} \cmidrule{26-26}
 & 0.3 & & 0.5 & & 0.7 & & 0.9  & & & & & & GT=98.24 & &  0.3 & & 0.5 & & 0.7 & & 0.9 & & & & GT=77.10  \\ 
\midrule
Ours w/o individual usage &  61.45  & &  43.22  & & 29.71 & & 20.05 & & 0.2652 & & 53.48 & & 94.96 & & 26.48 & & 14.29 & & 11.90 & & 9.81 & & 0.1264 & & 50.07 \\
Ours w/o weighted unification & 61.76 & & 45.28 & & 30.22 & & 20.51 & & 0.2795 & & 56.27 & & 95.07  & & 29.57 & & 18.22 & & 13.76 & & 10.80 & & 0.1501 & & 56.77 \\
\textbf{Ours} (completed model) & \bf 65.34 & & \bf 49.01 & & \bf 35.87 & & \bf 23.61 & & \bf 0.3186 & & \bf 68.23 & & \bf 97.19 & & \bf 35.00 & & \bf 23.12 & & \bf 16.34 & & \bf 13.40 & & \bf 0.1847 & & \bf 71.13\\
\midrule
Johnson+~\cite{johnson2018image} & 59.75 & & 42.53 & & 29.23 & & 19.89 & & 0.2532 & & 51.20 & & 94.82
& & 28.13 & & 17.17 & & 12.30 & & 10.47 & & 0.1485 & & 52.28\\
\multicolumn{26}{c}{\dotfill} \\
Zhang+~\cite{han2017stackgan} & 37.81 & & 20.50 & & 10.64 & & 7.76 & & 0.0824 & & 30.72 & & 60.15
& & 18.38 & & 10.84 & & 8.11 & & 5.82 & & 0.0643 & & 40.07 \\
Xu+~\cite{Tao18attngan} & 21.39 & & 10.71 & & 8.15 & & 5.83 & & 0.0671 & & 31.97 & & 52.76
& & 16.02 & & 9.33 & & 7.66 & & 5.15 & & 0.0579 & & 36.82 \\
\midrule[0.3pt]
\end{tabular}
}
\vspace*{-2\baselineskip}
\end{table}

\begin{table}[tb]
\begin{minipage}[t]{0.5\textwidth}
\centering
\caption{Comparison using caption generation metrics on COCO-stuff (larger is better; the best in \textcolor[rgb]{0,0,1}{blue}). Scores inside the parentheses indicate those reported in~\cite{hong2018inferring}.
}
\label{tab:caption_generation}
\resizebox{1\linewidth}{!}{
\begin{tabular}{l|@{\hspace*{0.35em}}c@{\hspace*{0.35em}}c@{\hspace*{0.35em}}c@{\hspace*{0.35em}}c@{\hspace*{0.35em}}c@{\hspace*{0.35em}}c@{\hspace*{0.35em}}c@{\hspace*{0.35em}}c@{\hspace*{0.35em}}c@{\hspace*{0.35em}}c@{\hspace*{0.35em}}c@{\hspace*{0.35em}}c@{\hspace*{0.35em}}c@{\hspace*{0.35em}}c}
\midrule[0.1pt]
Method & $BLEU-1$ & & $BLEU-2$ & & $BLEU-3$ & & $BLEU-4$ & & $METEOR$ & & $CIDEr$ \\
 \midrule
\textbf{Ours} & \textcolor[rgb]{0,0,1}{0.561} & & \textcolor[rgb]{0,0,1}{0.352} & & \textcolor[rgb]{0,0,1}{0.217} & & \textcolor[rgb]{0,0,1}{0.139} & & \textcolor[rgb]{0,0,1}{0.157} & & 0.325 \\
\midrule
Johnson+~\cite{johnson2018image} & 0.531 & & 0.321 & & 0.183 & & 0.107 & & 0.141 & & 0.238 \\
Hong+~\cite{hong2018inferring} & (0.541) & & (0.332) & & (0.199) & & (0.122) & & (0.154) & & (\textcolor[rgb]{0,0,1}{0.367})  \\
\multicolumn{12}{c}{\dotfill} \\
Zhang+~\cite{han2017stackgan} & 0.417 & & 0.214 & & 0.111 & & 0.062 & & 0.095 & & 0.078 \\
Xu+~\cite{Tao18attngan} & 0.450 & & 0.251 & & 0.157 & & 0.087 & & 0.105 & & 0.251 \\
\midrule[0.1pt]
GT & 0.627 & & 0.434 & & 0.287 & & 0.191 & & 0.191 & & 0.367 \\
 & (0.678) & & (0.496) & & (0.349) & & (0.243) & & (0.228) & & (0.802) \\
\midrule[0.1pt]
\end{tabular}
}
\vspace*{-1.5\baselineskip}

\end{minipage}
\hspace*{0.05\textwidth}
\begin{minipage}[t]{0.45\textwidth}
\centering
\caption{Comparison using diversity score~\cite{zhang2018perceptual} (the best in \textcolor[rgb]{0,0,1}{blue}; the second best in \textcolor[rgb]{1,0,0}{red}).
Scores are inside the parentheses indicates those in the original papers.
}
\label{tab:diversity_score}
\resizebox{0.9\linewidth}{!}{
\begin{tabular}{l|@{\hspace*{0.35em}}c@{\hspace*{0.35em}}c@{\hspace*{0.35em}}c}
\midrule[0.1pt]
Method & \textbf{COCO-stuff}~\cite{caesar2018coco} & & \textbf{GENOME}~\cite{johnson2015image} \\
 \midrule
\textbf{Ours} ($64 \times 64$) & 0.36$\pm$0.10 & & 0.39$\pm$0.09\\ 
\textbf{Ours} ($128 \times 128$) & 0.45$\pm$0.12 & & 0.49$\pm$0.07 \\
\textbf{Ours} ($256 \times 256$) & \textcolor[rgb]{1,0,0}{0.52$\pm$0.09} & & \textcolor[rgb]{0,0,1}{0.56$\pm$0.06} \\
\midrule
Johnson+~\cite{johnson2018image} & 0.29$\pm$0.10 & & 0.31$\pm$0.08 \\
Ashual+~\cite{ashual2019scenegeneration} & \textcolor[rgb]{0,0,1}{(0.67$\pm$0.05)} & & --- \\
Zhao+~\cite{zhaobo2019layout2im} & (0.15$\pm$0.06) & & (0.17$\pm$0.09) \\
Sun+~\cite{sun2019image} & (0.40$\pm$0.09) & & \textcolor[rgb]{1,0,0}{(0.43$\pm$0.09)} \\
\midrule[0.1pt]
\end{tabular}
}
\vspace*{-1.5\baselineskip}
\end{minipage}
\vspace*{-0.75\baselineskip}
\end{table}

Next, we evaluated how the scene structure given in input text was preserved in generated images using $R@\tau$ (we changed $\tau$ from $0.3$ to $0.9$ by $0.2$), $rIoU$, $RS$, and $coverage$, see Table~\ref{tab:quantitative_layout}.
We remark that we computed $RS$ only for COCO-stuff because COCO-stuff has geometrical relations only.
For Zhang+\cite{han2017stackgan} and Xu+\cite{Tao18attngan}, we employed Faster-RCNN~\cite{ren15fasterrcnn} to estimate their predicted BBs of entities where we set
the number of generated BBs to be the number of entities in an image.
We note that the number of predicted BBs by ours or Johnson+\cite{johnson2018image} was always the same with the number of entities in an image.

Table~\ref{tab:quantitative_layout} shows that our method performs best, indicating that our predicted BBs more precisely agree with those in relation (location and size) of entities given in texts than the compared methods.
To be more specific, 
$rIoU$'s in Table~\ref{tab:quantitative_layout} show that our predicted BBs more successfully retain the relations of entities than the other methods.
This observation is also supported by $RS$ on COCO-stuff.
Moreover, our method outperforms the others in $coverage$ and achieves comparable levels with the ground-truth BBs.
These indicate that our visual-relation layout is well-structured. 
Our method thus has even better ability of rendering more realistic images with multiple entities since the faithful scene structure and more BB coverage (i.e., entity information) are achieved.
Note that the observation that the $coverage$'s on COCO-stuff are better than those on GENOME explains the reason why generated images on COCO-stuff are better in $IS$ and $FID$ than those on GENOME.

Next, we use the image caption task to evaluate how the generated image is relevant to its input text.
We follow~\cite{hong2018inferring} to report scores on COCO-stuff~\cite{caesar2018coco}, see Table~\ref{tab:caption_generation}.
Note that we evaluated on COCO-stuff only since the pre-trained image caption model on GENOME is not available.
We also note that all the scores on the ground-truth dataset in~\cite{hong2018inferring} are higher than our re-computation.
Table~\ref{tab:caption_generation} shows that our method outperforms the others~\cite{johnson2018image,han2017stackgan,Tao18attngan,hong2018inferring} on $BLEU$, $METEOR$ and comparable to~\cite{hong2018inferring} on $CIDEr$. 
We thus conclude that our method performs more consistently with input texts than the others.

Finally, we show the diversity score of generated images in Table~\ref{tab:diversity_score}.
Overall, our scores are higher than Johnson+\cite{johnson2018image}, Zhao+\cite{zhaobo2019layout2im}, and Sun+\cite{sun2019image} on both COCO-stuff and GENOME, and comparable to Ashual+\cite{ashual2019scenegeneration} on COCO-stuff. Moreover, along with our stacking-GANs, our scores become better and better.
These scores also support the efficacy of our method.

We note that we observed that the number of (trainable) parameters in our model is about 41M which is comparable with Johnson+\cite{johnson2018image} (28M), Zhang+\cite{han2017stackgan} (57M), and Xu+\cite{Tao18attngan} (23M), and significantly smaller than Ashual\cite{ashual2019scenegeneration} (191M).

\subsection{Ablation study}

\begin{figure}[tb]
	\centering
	\includegraphics[width=0.9\linewidth]{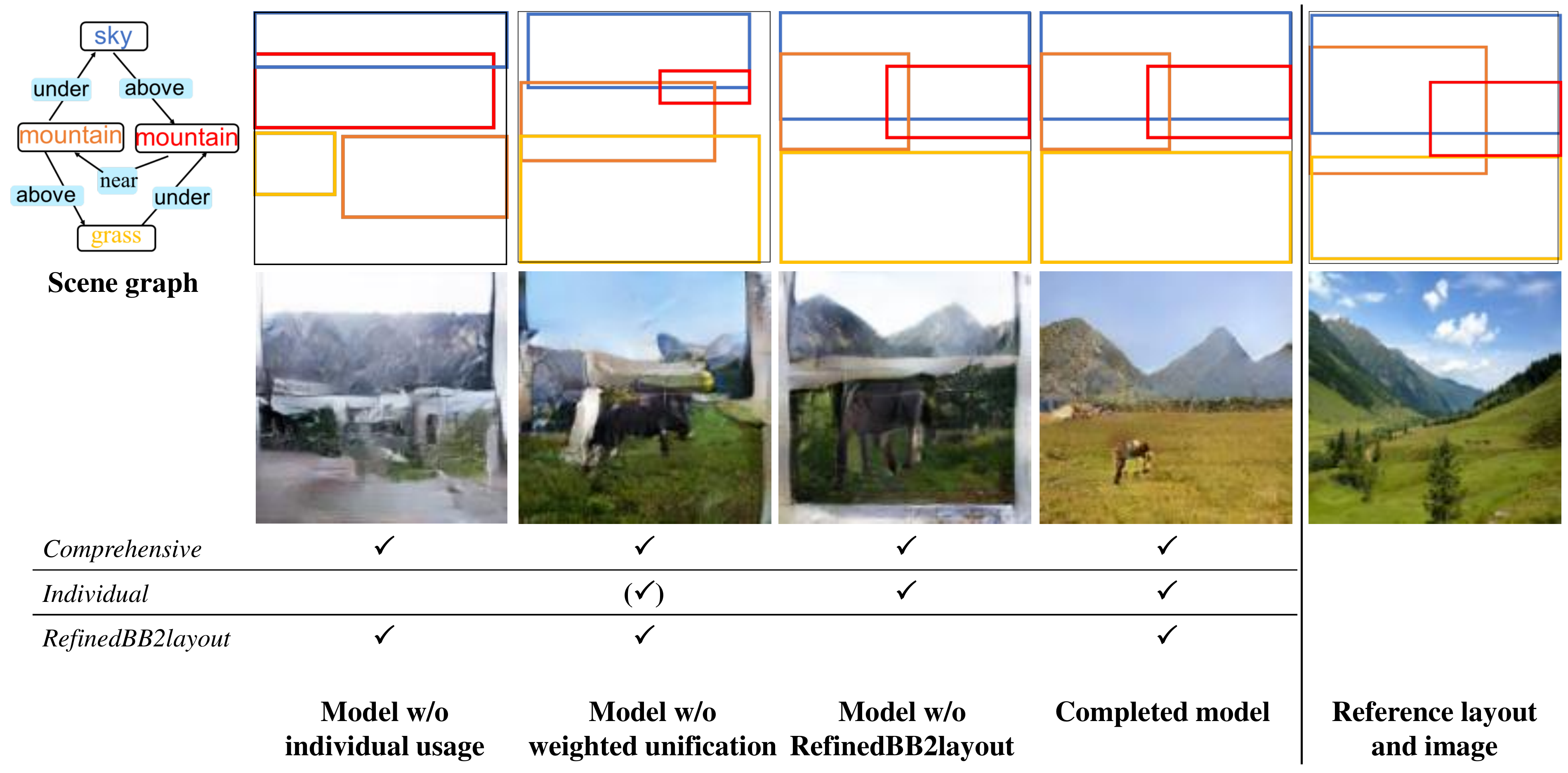}
    \vspace*{-0.5\baselineskip}
	\caption{Example of layouts and generated images by the ablation models. For each model, the 1st row shows the layout, the 2nd row shows the generated image. 
	All images are at $256 \times 256$ resolution.} 
	\label{fig:visualization_ablation}
	\vspace*{-1.5\baselineskip}
\end{figure}

We evaluated the plausibility of employing the visual-relation layout module, see the first block of Tables~\ref{tab:quantitative_IS_FID} and \ref{tab:quantitative_layout}: ours w/o individual usage denotes the model dropping the individual usage subnet; ours w/o weighted unification denotes the replacement of refining BBs with just averaging in the individual usage subnet; ours w/o refinedBB2layout denotes the replacement by just putting all entity layouts together in constructing the visual-relation layout. 
Fig.~\ref{fig:visualization_ablation} illustrates a typical output example of the ablation models.
We note that model w/o comprehensive usage is not applicable since all the other subnets in our visual-relation layout module need the output by the comprehensive usage subnet.

The 4th and 5th rows of Tables~\ref{tab:quantitative_IS_FID} and~\ref{tab:quantitative_layout} confirm the importance of the individual usage subnet. We also see the necessity of our learnable weights in refining BBs
because model w/o weighted unification performs better than model w/o individual usage.
We may conclude that the relation-unit prediction and the weighted unification together bring gain on our performance.

\begin{figure}[tb]
	\centering
	\includegraphics[width=0.8 \linewidth]{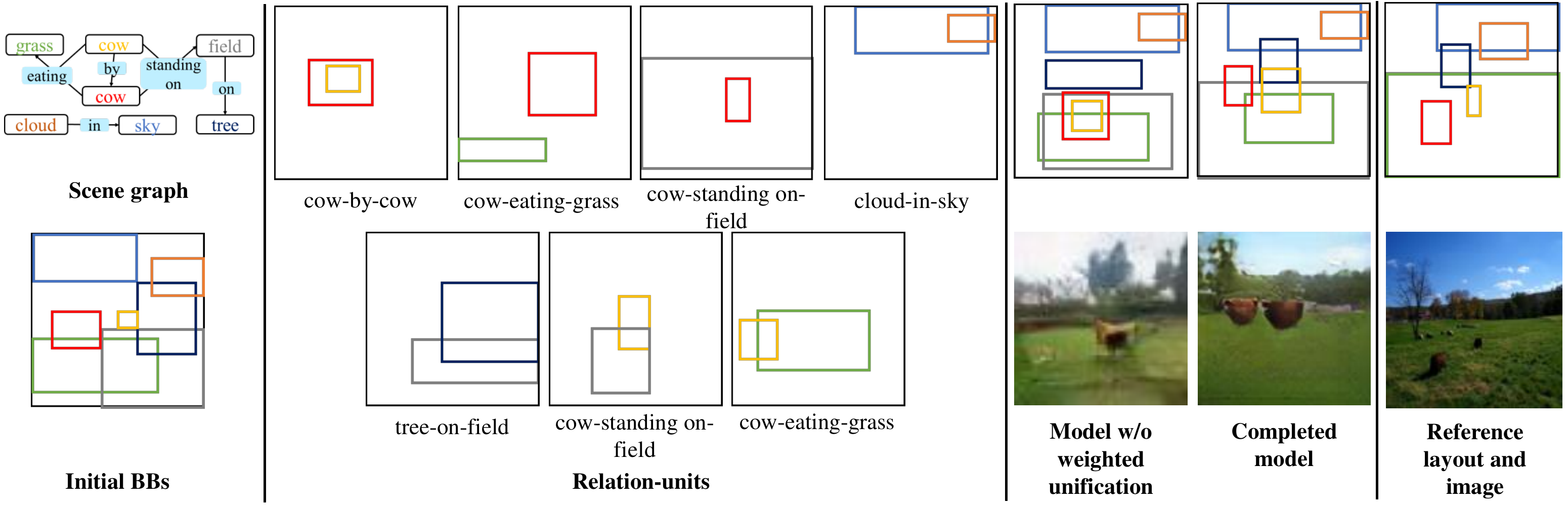}
    \vspace*{-0.8\baselineskip}
	\caption{Example of relation-units in the individual usage subnet; layouts and generated images by model w/o weighted unification and completed model.} 
	\label{fig:step-by-step}
	\vspace*{-0.75\baselineskip}
\end{figure}

From Fig.~\ref{fig:visualization_ablation}, we visually observe that the layout by the model w/o individual usage does not successfully reflect relations.  
This observation is applicable to the model w/o weighted unification as well.
As a result, both the models generated images in poorer quality than our complete model.
The relation-units are in diversity: entity BBs can be various in size and location because of multiple relations (see Fig.~\ref{fig:step-by-step}, for example), and thus
simply averaging BBs corresponding to the same entity does not successfully retain the relations among entities.
Therefore, the individual usage of relations in addition to the comprehensive usage is important for more consistent layout with input text.

The 6th row in Table~\ref{tab:quantitative_IS_FID} shows the significance of the refinedBB2layout.
Complex descriptions with many entities and relations tend to produce overlapped BBs.
The model w/o refinedBB2layout cannot necessarily produce all the entities in the layout, generating poor images.

We also evaluated the necessity of each term of the loss function through comparing our completed model with models dropping one term each: model w/o $\mathcal{L}_{\rm pix}$, model w/o $\mathcal{L}_{\rm context}$, and model w/o $\mathcal{L}_{\rm adv}$ (we dropped each term in the loss function (Sec.~\ref{loss function}) except for stacking-GANs).
From the 2nd block of Table~\ref{tab:quantitative_IS_FID}, we see that the absence of any term degrades the quality of generated images. 
This indicates that all the loss terms indeed contribute to performance.

\begin{figure}[tb]
	\centering
	\includegraphics[width=0.8\linewidth]{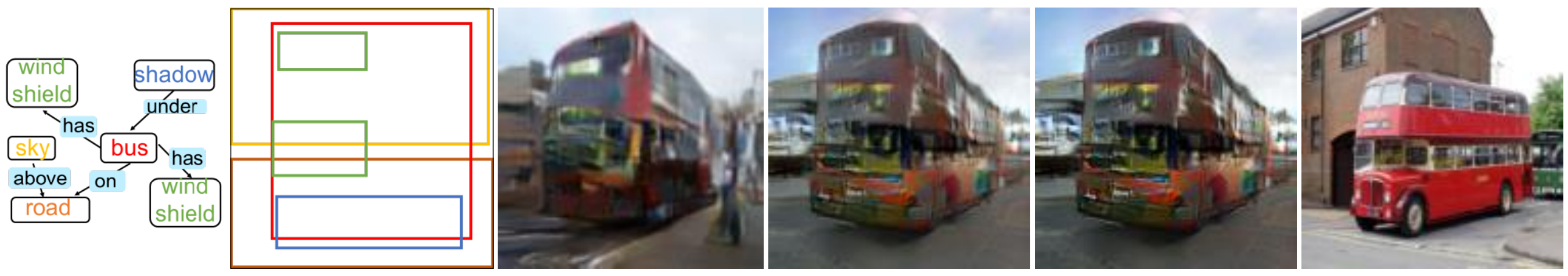}
    \vspace*{-0.8\baselineskip}
	\caption{Example of output along with the stacking-GANs. From left to right, scene graph, visual-relation layout, the outputs at $64 \times 64$, $128 \times 128$, $256 \times 256$ resolutions, and the reference image.} 
	\label{fig:stacking_example}
	\vspace*{-1.5\baselineskip}
\end{figure}

Finally, we see that along with the stacking of GANs, our method progressively generates better images 
in terms of $IS$ and $FID$ (Table~\ref{tab:quantitative_IS_FID}). 
We observe that at $64 \times 64$ resolution, generated images tend to be blurred and lose some details while
the details of images are improved as the resolution becomes higher (the best result is obtained at $256 \times 256$ resolution) (see Fig.~\ref{fig:stacking_example} as an example).
We also confirmed that the visual-relation layouts of generated images at any resolutions are the same and highly consistent with texts.

When we replaced CRN in~\cite{johnson2018image} with our stacking-GANs for $128\times 128$ and $256\times 256$ resolutions to factor out the influence of image generators,
we see that the improvement of~\cite{johnson2018image} on $IS$ and $FID$ along the resolution is worse than that of our model (the 10th and the 11th rows of Table~\ref{tab:quantitative_IS_FID}).
This indicates that better layout significantly improves the performance of the final image generation and also confirms clearer contribution of our proposed visual-relation layout module.

\section{Conclusion}
We proposed a GAN-based end-to-end network for text-to-image generation where 
relations between entities are comprehensively and individually used to infer a visual-relation layout.
We also conditioned the stacking-GANs on the visual-relation layout to generate high-resolution images.
Our layout preserves the scene structure more precisely than the layout by SOTAs.
Experimental results on two public datasets demonstrate the effectiveness of our method.

\bibliographystyle{splncs}
\bibliography{references}

\end{document}